\documentclass[twocolumn]{article}
\usepackage[dvips]{graphicx}
\usepackage{latexsym,amssymb}
\def\OO{\mathcal O} \def\RR{\mathcal R} \def\UU{\mathcal U} \def\PP{\mathcal P}
\def\NNN{\mathbb N} \def\QQQ{\mathbb Q} \def\BBB{\mathbb B}
\def\E{\mathbf{E}}

\begin{document}

\title{\vspace{-3ex}\normalsize\sc Technical Report \hfill IDSIA-10-06
\vskip 2mm\bf\Large\hrule height5pt \vskip 6mm
A Formal Measure of Machine Intelligence
\vskip 6mm \hrule height2pt}
\author{{\bf Shane Legg} and {\bf Marcus Hutter}\\[3mm]
\normalsize IDSIA, Galleria 2, CH-6928\ Manno-Lugano, Switzerland\\
\normalsize \{shane,marcus\}@idsia.ch \hspace{9ex} http://www.idsia.ch/ }
\date{14 April 2006}
\maketitle

\begin{abstract}
A fundamental problem in artificial intelligence is that nobody really
knows what intelligence is.  The problem is especially acute when we
need to consider artificial systems which are significantly different
to humans.  In this paper we approach this problem in the following
way: We take a number of well known informal definitions of human
intelligence that have been given by experts, and extract their
essential features.  These are then mathematically formalised to
produce a general measure of intelligence for arbitrary machines.  We
believe that this measure formally captures the concept of machine
intelligence in the broadest reasonable sense.
\end{abstract}

\section{Introduction}

Most of us think that we recognise intelligence when we see it, but we
are not really sure how to precisely define or measure it.  We
informally judge the intelligence of others by relying on our past
experiences in dealing with people.  Naturally, this naive approach is
highly subjective and imprecise.  A more principled approach would be
to use one of the many standard intelligence tests that are available.
Contrary to popular wisdom, these tests, when correctly applied by a
professional, deliver statistically consistent results and have
considerable power to predict the future performance of individuals in
many mentally demanding tasks.  However, while these tests work well
for humans, if we wish to measure the intelligence of other things,
perhaps of a monkey or a new machine learning algorithm, they are
clearly inappropriate.

One response to this problem might be to develop specific kinds of
tests for specific kinds of entities; just as intelligence tests for
children differ to intelligence tests for adults.  While this works
well when testing humans of different ages, it comes undone when we
need to measure the intelligence of entities which are profoundly
different to each other in terms of their cognitive capacities, speed,
senses, environments in which they operate, and so on.  To measure the
intelligence of such diverse systems in a meaningful way we must step
back from the specifics of particular systems and establish the
underlying fundamentals of what it is that we are really trying to
measure.  That is, we need to establish a notion of intelligence that
goes beyond the specifics of particular kinds of systems.

The difficulty of doing this is readily apparent.  Consider, for
example, the memory and numerical computation tasks that appear in
some intelligence tests and which were once regarded as defining
hallmarks of human intelligence.  We now know that these tasks are
absolutely trivial for a machine and thus do not test the machine's
intelligence.  Indeed even the mentally demanding task of playing
chess has been largely reduced to brute force search.  As technology
advances, our concept of what intelligence is continues to evolve with
it.

How then are we to develop a concept of intelligence that is
applicable to all kinds of systems?  Any proposed definition must
encompass the essence of human intelligence, as well as other
possibilities, in a consistent way.  It should not be limited to any
particular set of senses, environments or goals, nor should it be
limited to any specific kind of hardware, such as silicon or
biological neurons.  It should be based on principles which are
sufficiently fundamental so as to be unlikely to alter over time.
Furthermore, the intelligence measure should ideally be formally
expressed, objective, and practically realisable.

This paper approaches this problem in the following way.  In
\emph{Section \ref{sec:infint}} we consider a range of definitions of
human intelligence that have been put forward by well known
psychologists.  From these we extract the most common and essential
features and use them to create an informal definition of
intelligence.  \emph{Section \ref{sec:aef}} then introduces the
framework which we use to construct our formal measure of
intelligence.  This framework is formally defined in \emph{Section
\ref{sec:formframe}}.  In \emph{Section \ref{sec:ior}} we use our
developed formalism to produce a formal definition of intelligence.
\emph{Section \ref{sec:conc}} closes with a short summary.

A preliminary sketch of the ideas in this paper appeared in the poster
\cite{Legg:05iors}.  It can be shown that the intelligence measure
presented here is in fact a variant of the Intelligence Order Relation
that appears in the theory of AIXI, the provably optimal universal
agent \cite{Hutter:04uaibook}.  A long journal version of this paper
is being written in which we give the proposed measure of machine
intelligence and its relation to other such tests a much more
comprehensive treatment.

Naturally, we expect such a bold initiative to be met with resistance.
However, we hope that the reader will appreciate the value of our
approach: With a formally precise definition put forward we aim to
better our understanding of what is a notoriously subjective and
slippery concept.

\section{The concept of intelligence}\label{sec:infint}

Although definitions of human intelligence given by experts in the
field vary, most of their views cluster around a few common
perspectives.  Perhaps the most common perspective, roughly stated, is
to think of intelligence as being the ability to successfully operate
in uncertain environments by learning and adapting based on
experience.  The following often quoted definitions, which can be
found in \cite{Sternberg:00}, \cite{Wechsler:58}, \cite{Bingham:37}
and \cite{Gottfredson:97msoi}, all express this notion of intelligence
but with different emphasis in each case:

\begin{itemize}

\item ``The capacity to learn or to profit by experience.''
\mbox{--~W. F. Dearborn}

\item ``Ability to adapt oneself adequately to relatively new situations in
life.''
--~R. Pinter

\item ``A person possesses intelligence insofar as he has learned, or can
learn, to adjust himself to his environment.''
--~S. S. Colvin

\item ``We shall use the term `intelligence' to mean the ability of an
organism to solve new problems\ldots.''
\mbox{--~W. V. Bingham}

\item ``A global concept that involves an individual's ability to
act purposefully, think rationally, and deal effectively with
the environment.''
\mbox{-- D. Wechsler}

\item ``Intelligence is a very general mental capability that, among
other things, involves the ability to reason, plan, solve problems,
think abstractly, comprehend complex ideas, learn quickly and learn
from experience.''
\mbox{--~L. S. Gottfredson and 52 expert signatories}

\end{itemize}

These definitions have certain common features; in some cases they are
explicitly stated, while in others they are more implicit.  Perhaps
the most elementary feature is that intelligence is seen as a property
of an entity which is interacting with an external environment,
problem or situation.  Indeed this much is common to practically all
proposed definitions of intelligence.  As we will be referring back to
these concepts regularly, we will refer to the entity whose
intelligence is in question as the \emph{agent}, and the external
environment, problem or situation that it faces as the
\emph{environment}.  An environment could be a large complex world in
which the agent exists, similar to the usual meaning, or something as
narrow as a game of tic-tac-toe.

The second common feature of these definitions is that an agent's
intelligence is related to its ability to succeed in an environment.
This implies that the agent has some kind of an objective.  Perhaps we
could consider an agent intelligent, in an abstract sense, without
having any objective.  However without any objective what so ever, the
agent's intelligence would have no observable consequences.
Intelligence then, at least the concrete kind that interests us, comes
into effect when an agent has an objective to apply its intelligence
to.  Here we will refer to this as its \emph{goal}.

The emphasis on learning, adaption and experience in these definitions
implies that the environment is not fully known to the agent and may
contain surprises and new situations which could not have been
anticipated in advance.  Thus intelligence is not the ability to deal
with one fixed and known environment, but rather the ability to deal
with some range of possibilities which cannot be wholly anticipated.
This means that an intelligent agent may not be the best possible in
any specific environment, particularly before it has had sufficient
time to learn.  What is important is that the agent is able to learn
and adapt so as to perform well over a wide range of specific
environments.

Although there is a great deal more to this topic than we have
presented here, the above brief analysis gives us the necessary
building blocks for our informal working definition of intelligence:
\begin{quote}
\emph{Intelligence measures an agent's ability to achieve goals in a
wide range of environments.}
\end{quote}

We realise that some researchers who study intelligence will take
issue with this definition.  Given the diversity of views on the
nature of intelligence, a debate which is still being fought, this is
unavoidable.  Nevertheless, we are confident that our proposed
informal working definition is fairly mainstream.  We also believe
that our definition captures what we are interested in achieving in
machines: A very general and flexible capacity to succeed when faced
with a wide range of problems and situations.  Even those who
subscribe to different perspectives on the nature and correct
definition of intelligence will surely agree that this is a central
objective for anyone wishing to extend the power and usefulness of
machines.  It is also a definition that can be successfully
formalised.

\section{The agent-environment framework}\label{sec:aef}

In the previous section we identified three essential components for
our model of intelligence: An agent, an environment, and a goal.
Clearly, the agent and the environment must be able to interact with
each other; specifically, the agent needs to be able to send signals
to the environment and also receive signals being sent from the
environment.  Similarly the environment must be able to receive and
send signals to the agent.  In our terminology we will adopt the
agent's perspective on these communications and refer to the signals
from the agent to the environment as \emph{actions}, and the signals
from the environment as \emph{perceptions}.

What is missing from this setup is the goal.  As discussed in the
previous section, our definition of an agent's intelligence requires
there to be some kind of goal for the agent to try to achieve.  This
implies that the agent somehow knows what the goal is.  One
possibility would be for the goal to be known in advance and for this
knowledge to be built into the agent.  The problem with this however
is that it limits each agent to just one goal.  We need to allow
agents which are more flexible than this.

If the goal is not known in advance, the other alternative is to
somehow inform the agent of what the goal is.  For humans this is
easily done using language.  In general however, the possession of a
sufficiently high level of language is too strong an assumption to
make about the agent.  Indeed, even for something as intelligent as a
dog or a cat, direct explanation will obviously not work.

Fortunately there is another possibility.  We can define an additional
communication channel with the simplest possible semantics: A signal
that indicates how good the agent's current situation is.  We will
call this signal the \emph{reward}.  The agent's goal is then simply
to maximise the amount of reward it receives, so in a sense its goal
is fixed.  This is not limiting though as we have not said anything
about what causes different levels of reward to occur.  In a complex
setting the agent might be rewarded for winning a game or solving a
difficult puzzle.  From a broad perspective then, the goal is
flexible.  If the agent is to succeed in its environment, that is,
receive a lot of reward, it must learn about the structure of the
environment and in particular what it needs to do in order to get
reward.

\begin{figure}[t]
\centerline{\includegraphics[width=0.77\columnwidth]{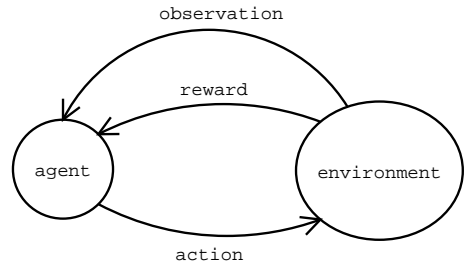}}
\caption{\label{agent-env}The agent and the environment interact by
sending action, observation and reward signals to each other.}
\end{figure}

Not surprisingly, this is exactly the way in which we condition an
animal to achieve a goal: by selectively rewarding certain behaviours.
In a narrow sense the animal's goal is fixed, perhaps to get more
treats to eat, but in a broader sense this may require doing a trick
or solving a puzzle.

In our framework we will include the reward signal as a part of the
perception generated by the environment.  The perceptions also contain
a non-reward part, which we will refer to as \emph{observations}.
This now gives us the complete system of interacting agent and
environment in Figure~\ref{agent-env}.  The goal, in the broad
flexible sense, is implicitly defined by the environment as this is
what defines when rewards are generated.  Thus in this framework, to
test an agent in any given way, it is sufficient to fully define the
environment.

In artificial intelligence, this framework is used in the area of
reinforcement learning \cite{Sutton:98}.  By appropriately renaming
things, it also describes the controller-plant framework used in
control theory.  It is a widely used and very general structure that
can describe seemingly any kind of learning or control problem.  The
interesting point for us is that this type of framework follows
naturally from our informal definition of intelligence.  The only
difficulty was how to deal with the notion of success, or profit.
This requires the existence of some kind of objective or goal, and the
most flexible and elegant way to bring this into our framework is by
using a simple reward signal.

\section{A formal framework for intelligence} \label{sec:formframe}

Having made the basic framework explicit, we can now formalise things.
See \cite{Hutter:04uaibook} for a more complete technical description
along with many more example agents and environments.

The agent sends information to the environment by sending
\emph{symbols} from some finite set, for example, $\AA := \{ left,
right, forwards, backwards \}$.  We will call this set the
\emph{action space} and denote it by $\AA$.  Similarly, the
environment sends signals to the agent with symbols from a finite set
called the \emph{perception space}, which we will denote $\PP$.  The
\emph{reward space}, denoted by $\RR$, will always be a finite subset
of the rational unit interval $[0,1] \cap \QQQ$.  Every perception
consists of two separate parts; an observation and a reward.  For
example, we might have $\PP := \{ (cold, 0.0), (warm, 1.0), (hot,
0.3), (roasting, 0.0) \}$.

To denote symbols being sent we will use the lower case variable names
$a$, $o$ and $r$ for actions, observations and rewards respectively.
We will also index these in the order in which they occur, thus $a_1$
is the agent's first action, $a_2$ is the second action and so on.
The agent and the environment will take turns at sending symbols,
starting with the environment.  This produces a history of
observations, rewards and actions which we will denote by, $o_1 r_1
a_1 o_2 r_2 a_2 o_3 r_3 a_3 o_4 \ldots$.  Our restriction to finite
action and perception spaces is deliberate as an agent should not be
able to receive or generate information without bound in a single
cycle in time.  Of course, the action and perception spaces can still
be extremely large, if required.

Formally, the agent is a function, denoted by $\pi$, which takes the
current history as input and chooses the next action as output.  A
convenient way of representing the agent is as a probability measure
over actions conditioned on the current history.  Thus $\pi( a_3 | o_1
r_1 a_1 o_2 r_2 )$ is the probability of action $a_3$ in the third
cycle, given that the current history is $o_1 r_1 a_1 o_2 r_2$.  A
deterministic agent is simply one that always assigns a probability of
1 to some action for any given history.  How the agent produces the
distribution over actions for any given history is left completely
open.  Of course in artificial intelligence the agent will be a
machine and so $\pi$ will be a computable function.

The environment, denoted $\mu$, is defined in a similar way.
Specifically, for any $k \in \NNN$ the probability of $o_k r_k$, given
the current history $o_1 r_1 a_1 \ldots o_{k-1} r_{k-1} a_{k-1}$, is
$\mu( o_k r_k | o_1 r_1 a_1 \ldots o_{k-1} r_{k-1} a_{k-1} )$.  For
the moment we will not place any further restrictions on the
environment.

Our next task is to formalise the idea of ``profit'' or ``success''
for an agent.  Informally, we know that the agent must try to maximise
the amount of reward it receives, however this could mean several
different things.

\vspace{1em}

\noindent {\bf Example.}  Define the reward space $\RR := \{ 0, 1 \}$, an action
space $\AA := \{ 0, 1 \}$ and an observation space that just contains
the null string, $\OO := \{ \varepsilon \}$.  Now define a simple
environment,
\[
\mu( r_k | o_1 \ldots a_{k-1} ) := 1 - | r_k - a_{k-1} |.
\]
As the agent always get a reward equal to its action, the optimal
agent for this environment is clearly $\pi_{opt} ( a_k | o_1 \ldots
r_k ) := a_k$.  Consider now two other agents for this environment,
$\pi_1 ( a_k | o_1 \ldots r_k ) = \frac{1}{2}$ and
\begin{displaymath}
\pi_2( a_k | o_1 \ldots r_k ) := \left\{
\begin{array}{ll}
1 & \mathrm{for\ } a_k = 0 \land k \leq 100,\\
1 & \mathrm{for\ } a_k = 1 \land 100 < k \leq 5000,\\
\frac{1}{2} & \mathrm{for\ }  5000 < k,\\
0 & \mathrm{otherwise}.
\end{array} \right.
\end{displaymath}

For $1 \leq k \leq 100$ the expected reward per cycle for $\pi_1$ is
higher than it is for $\pi_2$.  Thus in the short term $\pi_1$ is the
most successful.  On the other hand, for $100 < k \leq 5000$, $\pi_2$
has switched to the optimal strategy of always guessing that 1 head
will be thrown, while $\pi_1$ has not.  Thus in the medium term
$\pi_2$ is more successful.  Finally, for $k > 5000$, both agents use
random actions and thus in the limit they are equally successful.

Which is the better agent?  If you want to maximise short term
rewards, it is agent $\pi_1$.  If you want to maximise medium term
rewards, then it is agent $\pi_2$.  And if you only care about the
long run, both agents are equally successful.  Which agent you prefer
depends on your temporal preferences, something which is currently
outside of our formulation.

The standard way of formalising this in reinforcement learning is to
assume that the value of rewards decay geometrically into the future
at a rate given by a discount parameter $\gamma \in (0,1)$, that is,
\begin{equation}\label{eqn:disval}
V^{\pi}_{\mu}(\gamma) := \: \frac{1}{\Gamma} \E \left(
\sum_{i=1}^\infty \gamma^i r_i \right)
\end{equation}
where $r_i$ is the reward in cycle $i$ of a given history, the
normalising constant is $\Gamma := \sum_{i=1}^\infty \gamma^i$, and
the expected value is taken over all histories of $\pi$ and $\mu$
interacting.  By increasing $\gamma$ towards 1 we weight long term
rewards more heavily, conversely by reducing it we balance the
weighting towards short term rewards.

Of course this has not actually answered the question of how to weight
near term rewards versus longer term rewards.  Rather it has simply
expressed this weighting as a parameter.  While that is adequate for
some purposes, what we would like is a single test of intelligence for
machines, not a range of tests that vary according to some free
parameter.  That is, we would like the temporal preferences to be
included in the model, not external to it.

One possibility might be to use harmonic discounting, $\gamma_t :=
\frac{1}{t^2}$.  This has some nice properties, in particular the
agent needs to look forward into the future in a way that is
proportional to its current age \cite{Hutter:04uaibook}.  However an
even more elegant solution is possible.

If we look at the value function in Equation~\ref{eqn:disval}, we see
that geometric discounting plays two roles.  Firstly, it normalises
the total reward received which makes the sum finite, in this case
with a maximum value of 1.  Secondly, it weights the reward at
different points in the future which in effect defines a temporal
preference.  We can solve both of these problems, without needing an
external parameter, by simply requiring that the total reward returned
by the environment cannot exceed 1.  For a reward summable environment
$\mu$ we can now define the value function to be simply,
\begin{equation}\label{eqn:unival}
V^{\pi}_{\mu} := \: \E \left( \sum_{i=1}^\infty r_i \right)
\leq 1.
\end{equation}

One way of viewing this is that the rewards returned by the
environment now have the temporal preference factored in and thus we
do not need to add this.  The cost is that this is an additional
condition that we place on the environments.  Previously we required
that each reward signal was in a finite subset of $[0,1] \cap \QQQ$,
now we have the additional constraint that the sum is bounded.

It may seem that there is a philosophical problem here.  If an
environment $\mu$ is an artificial game, like chess, then it seems
fairly natural for $\mu$ to meet any requirements in its definition,
such as having a bounded reward sum.  However if we think of the
environment $\mu$ as being ``the universe'' in which the agent lives,
then it seems unreasonable to expect that it should be required to
respect such a bound.  The flaw in this argument is that a
``universe'' does not have any notion of reward for particular agents.

Strictly speaking, reward is an interpretation of the state of the
environment.  In humans this is built in, for example, the pain that
is experienced when you touch something hot.  In which case, maybe it
should really be a part of the agent rather than the environment?  If
we gave the agent complete control over rewards then our framework
would become meaningless: The perfect agent could simply give itself
constant maximum reward.  Indeed humans cannot easily do this either,
at least not without taking drugs designed to interfere with their
pleasure-pain mechanism.

Thus the most accurate framework would consist of an agent, an
environment and a separate goal system that interpreted the state of
the environment and rewarded the agent appropriately.  In such a set
up the bounded rewards restriction would be a part of the goal system
and thus the above philosophical problem does not occur.  However for
our current purposes it is seem sufficient just to fold this goal
mechanism into the environment and add an easily implemented
constraint to how the environment may generate rewards.

\section{A formal measure of intelligence}\label{sec:ior}

We have now formally defined the space of agents, how they interact
with each other, and how we measure the performance of an agent in any
specific environment.  Before we can put all this together into a
single performance measure, we firstly need to define what me mean by
``a wide range of environments.''

As our goal is to produce a measure of intelligence that is as broad
and encompassing as possible, the space of environments used in our
definition should be as large as possible.  Given that our environment
is a probability measure with a certain structure, an obvious
possibility would be to consider the space of all probability measures
of this form.  Unfortunately, this extremely broad class of
environments causes problems.  As the space of all probability
measures is uncountably infinite, we cannot list the members of this
set, nor can we always describe environments in a finite way.

The solution is to require the environmental measures to be
computable.  Not only is this necessary if we are to have an effective
measure of intelligence, it is also not all that restrictive.  There
are an infinite number of environments in this set, with no upper
bound on their complexity.  Furthermore, it is only the measure which
describes the environment that must be computable.  For example,
although a typical sequence of 1's and 0's generated by flipping a
coin is not computable, the probability measure which describes this
process is computable.  Thus, even environments which behave randomly
are included in our space of environments.  This appears to be the
largest reasonable space of environments.  Indeed, no physical system
has ever been shown to lie outside of this set.  If such a physical
system was found, it would overturn the Church-Turing thesis and alter
our view of the universe.

How can we combine the agent's performance over all these
environments?  As there are an infinite number of environments, we
cannot simply take a uniform distribution over them.  Mathematically,
we must weight some environments more highly than others.  If we
consider the agent's perspective on the problem, this question is the
same as asking: Given several different hypotheses which are
consistent with the data, which hypothesis should be considered the
most likely?  This is a frequently occurring problem in inductive
inference where we must employ a philosophical principle to decide
which hypothesis is the most likely.  The most successful approach is
to invoke the principle of Occam's razor: Given multiple hypotheses
which are consistent with the data, the simplest should be preferred.
This is generally considered the rational and intelligent thing to do.

Consider for example the following type of question which commonly
appears in intelligence tests.  There is a sequence such as 2, 4, 6,
8, and the test subject needs to predict the next number.  Of course
the pattern is immediately clear: The numbers are increasing by 2 each
time.  An intelligent person would easily identify this pattern and
predict the next digit to be 10.  However, the polynomial $2k^4 -20k^3
+70k^2 -98k +48$ is also consistent with the data, in which case the
next number in the sequence would be 58.  Why then do we consider the
first answer to be more likely?  It is because we use, perhaps
unconsciously, the principle of Occam's razor.  Furthermore, the fact
that the test defines this as the correct answer shows that it too
embodies the concept of Occam's razor.  Thus, although we don't
usually mention Occam's razor when defining intelligence, the ability
to effectively use Occam's razor is clearly a part of intelligent
behaviour.

Our formal measure of intelligence needs to reflect this.
Specifically, we need to test the agents in such a way that they are,
at least on average, rewarded for correctly applying Occam's razor.
Formally, this means that our a priori distribution over environments
should be weighted towards simpler environments.  The problem now
becomes: How should we measure the complexity of environments?

As each environment is computable, it can be represented by a program,
or more formally, a binary string $p \in \BBB^*$ on some prefix
universal Turing machine $\UU$.  Thus we can use Kolmogorov complexity
to measure the complexity of an environment $\mu \in E$,
\[
K( \mu ) := \min_{p \in \BBB^*} \big\{ |p| : \UU(p)
\mathrm{\ computes\ } \mu \big\}.
\]
This measure is independent of the choice of $\UU$ up to an additive
constant that is independent of $\mu$, thus, we simply pick one
universal Turing machine $\UU$ and fix it.  The correct way to turn
this into a prior distribution is by taking $2^{-K(\mu)}$.  This is
known as the algorithmic probability distribution and it has a number
of important properties, particularly in the context of universally
optimal learning agents.  See \cite{Li:97} or \cite{Hutter:04uaibook}
for an overview of Kolmogorov complex and universal prior
distributions.

Putting this all together, we can now define our formal measure of
intelligence for arbitrary systems.  Let $E$ be the space of all
programs that compute environmental measures of summable reward with
respect to a prefix universal Turing machine $\UU$, let $K$ be the
Kolmogorov complexity function.  The intelligence of an agent $\pi$ is
defined as,
\[
\Upsilon(\pi) :=  \sum_{\mu \in E} 2^{-K(\mu)} V^{\pi}_{\mu} = V^{\pi}_{\xi},
\]
where $\xi := \sum_{\mu \in E} 2^{-K(\mu)} \mu$ due to the linearity
of $V$.  $\xi$~is the Solomonoff-Levin universal a priori distribution
generalised to reactive environments.

\section{Properties of the intelligence measure}

To better understand the performance of this measure consider some
example agents.

\emph{A random agent.}  The agent with the lowest intelligence, at
least among those that are not actively trying to perform badly, would
be one that makes uniformly random actions.  We will call this
$\pi^\mathtt{rand}$.  In general such an agent will not be very
successful as it will fail to exploit any regularities in the
environment, no matter how simple they are.  It follows then that the
values of $V^{\pi^\mathtt{rand}}_\mu$ will typically be low compared
to other agents, and thus $\Upsilon (\pi^\mathtt{rand})$ will be low.

\emph{A very specialised agent.}  From the equation for $\Upsilon$, we
see that an agent could have very low intelligence but still perform
extremely well at a few very specific and complex tasks.  Consider,
for example, IBM's Deep Blue chess supercomputer, which we will
represent by $\pi^\mathtt{dblue}$.  When $\mu^\mathtt{chess}$
describes the game of chess,
$V^{\pi^\mathtt{dblue}}_{\mu^\mathtt{chess}}$ is very high.  However
$2^{-K(\mu^\mathtt{chess})}$ is small, and for $\mu \neq
\mu^\mathtt{chess}$ the value function will be low relative to other
agents as $\pi^\mathtt{dblue}$ only plays chess.  Therefore, the value
of $\Upsilon (\pi^\mathtt{dblue})$ will be very low.  Intuitively,
this is because Deep Blue is too inflexible and narrow to have general
intelligence.

\emph{A general but simple agent.}  Imagine an agent that does very
basic learning by building up a table of observation and action pairs
and keeping statistics on the rewards that follow.  Each time an
observation that has been seen before occurs, the agent takes the
action with highest estimated expected reward in the next cycle with
90\% probability, or a random action with 10\% probability.  We will
call this agent $\pi^\mathtt{basic}$.  It is immediately clear that
many environments, both complex and very simple, will have at least
some structure that such an agent would take advantage of.  Thus for
almost all $\mu$ we will have $V^{\pi^\mathtt{basic}}_\mu >
V^{\pi^\mathtt{rand}}_\mu$ and so $\Upsilon (\pi^\mathtt{basic}) >
\Upsilon (\pi^\mathtt{rand})$.  Intuitively, this is what we would
expect as $\pi^\mathtt{basic}$, while very simplistic, is surely more
intelligent than $\pi^\mathtt{rand}$.

\emph{A simple agent with more history.}  A natural extension of
$\pi^\mathtt{basic}$ is to use a longer history of actions,
observations and rewards in its internal table.  Let
$\pi^\mathtt{2back}$ be the agent that builds a table of statistics
for the expected reward conditioned on the last two actions, rewards
and observations.  It is immediately clear $\pi^\mathtt{2back}$ is a
generalisation of $\pi^\mathtt{basic}$ by definition and thus will
adapt to any regularity that $\pi^\mathtt{basic}$ can adapt to.  It
follows then that in general $V^{\pi^\mathtt{2back}}_\mu >
V^{\pi^\mathtt{basic}}_\mu$ and so $\Upsilon (\pi^\mathtt{2back}) >
\Upsilon (\pi^\mathtt{basic})$, as we would intuitively expect.

In a similar way agents of increasing complexity and adaptability can
be defined which will have still greater intelligence.  However with
more complex agents it is usually difficult to theoretically establish
whether one agent has more or less intelligence than another.
Nevertheless, it is hopefully clear from these simple examples that
the more flexible and powerful an agent is, the higher its machine
intelligence.

\emph{A human.}  For extremely simple environments, a human should be
able to identify their simple structure and exploit this to maximise
reward.  For more complex environments however it is hard to know how
well a human would perform without experimental results.

\emph{Super-human intelligence.}  It can be easily proven that the
theoretical AIXI agent \cite{Hutter:04uaibook} is the maximally
intelligent agent with respect to $\Upsilon$.  AIXI has been proven to
have many universal optimality properties, including being Pareto
optimal and self-optimising in any environment in which this is
possible for a general agent.  Thus it is clear that agents with very
high $\Upsilon$ must be extremely powerful.

In addition to sensibly ordering many simple learning agents, this
formal definition has many significant and desirable properties:

\emph{Valid}.  The most important property of a measure of
intelligence is that it does indeed measure ``intelligence''.  As
$\Upsilon$ formalises a mainstream informal definition, we believe
that it is valid measure.

\emph{Meaningful}. An agent with a high $\Upsilon$ value must perform
well over a very wide range of environments, in particular it must
perform well in almost all simple environments.  If such a agent
existed, it would clearly be very powerful and practically useful.  It
also sensibly orders the intelligence of simple learning agents.

\emph{Repeatable}. We can test an agent using the $\Upsilon$
repeatedly without problem.  This is because it is defined across all
well defined environments, not just a specific test subset which an
agent might adapt to.

\emph{Absolute}.  $\Upsilon$ gives us a single real absolute value,
unlike the pass-fail Turing test \cite{Turing:50}.  This is important
if we want to make distinctions between similar learning algorithms
that are not close to human level intelligence.

\emph{Wide range}. As we have seen, $\Upsilon$ can measure performance
from extremely simple agents right up to the super powerful AIXI
agent.  Other tests cannot hand such an enormus range.

\emph{General}.  The test is clearly non-specific to the
implementation of the agent as the inner workings of the agent is left
completely undefined.  It is also very general in terms of what senses
or actuators the agent might have as all information exchanged between
the agent and the environment takes place over basic Shannon like
communication channels.

\emph{Dynamic}.  One aspect of our test of intelligence is that it is,
in the terminology of intelligence testing, a highly dynamic test
\cite{Sternberg:02}.  Normally intelligence tests for humans only test
the ability to solve one-off problems.  There are no dynamic aspects
to the test where the test subject has to interact with something and
learn and adapt their behaviour accordingly.  This makes it very hard
to test things like the individual's ability to quickly pick up new
skills and adapt to new situations.  One way to overcome these
problems is to use more sophisticated dynamic tests.  In these tests
there is an active tester who constantly interacts with the test
subject, much like what happens in our formal intelligence measure.

\emph{Unbiased}.  The test is not weighted towards ability in certain
specific kinds of areas or problems, rather it is simply weighted
towards simpler environments no matter what they are.

\emph{Fundamental}.  The test is based on the theory of information,
Turing computation and complexity theory.  These are all fundamental
ideas which are likely to remain very stable over time irrespective of
changes in technology.

\emph{Formal}.  Unlike many tests of intelligence, $\Upsilon$ is
completely formally, mathematically, specified.

\emph{Objective}.  Unlike the Turing test which requires a panel of
judges to decide if an agent is intelligent or not, $\Upsilon$ is fee
of such subjectivity.

Our definition of intelligence also has some weaknesses.  One is the
fact that the environmental distribution $2^{-K(\mu)}$ that we have
used is invariant, up to a multiplicative constant, to changes in the
reference machine $\UU$.  While this affords us some protection, it
still means that the relative intelligence of agents can change if we
change our reference machine.  One approach to this problem might be
to limit the complexity of the reference machine, for example by
limiting its state-symbol complexity.  We expect that for highly
intelligent machines that can deal with a wide range of environments
of varying complexity, the effect of changing from one simple
reference machine to another will be minor.  For agents which are less
complex than the reference machine however, such a change could be
significant.

A theoretical problem is that our distribution over environments is
not computable.  While this is fine for a theoretical definition of
intelligence, it makes the measure impossible to directly implement.
The solution is to use a more tractable measure of complexity such as
Levin's $Kt$ complexity \cite{Levin:73search}, or Schmidhuber's Speed
prior \cite{Schmidhuber:02speed}.  Both of these consider the
complexity of an algorithm to be determined by both its description
length and running time.  Intuitively it also makes good sense,
because we would not usually consider a very short algorithm that
takes an enormous amount of time to compute, to be a particularly
simple one.

The only closely related work to ours is the C-Test
\cite{Hernandez:00btt}.  While our intelligence measure is fully
dynamic and interactive, the C-Test is a purely static sequence
prediction test similar to standard IQ tests for humans.  The C-Test
always ensures that each question has an unambiguous answer in the
sense that there is always one consistent hypothesis with
significantly lower complexity than the alternatives.  Perhaps this is
useful for some kinds of tests, but we believe that it is unrealistic
and limiting.  Like our intelligence test, the C-Test also has to deal
with the problem of the incomputability of Kolmogorov complexity.  By
using Levin's $Kt$ complexity, the C-Test was able to compute a number
of test problems which were used to test humans.  The ``compression
test''\cite{Mahoney:99} for machine intelligence is similarly
restricted to sequence prediction.  We consider the linguistic
complexity tests of Treister-Goren et.\ al.\ to be far too narrow.
The psychometric approach of Bringsjord and Schimanski is only
appropriate if the machine has a sufficiently human-like intelligence.

\section{Conclusions}\label{sec:conc}

Given the obvious significance of formal definitions of intelligence
for research, and calls for more direct measures of machine
intelligence to replace the problematic Turing test and other
imitation based tests \cite{Johnson:92}, very little work has been
done in this area.  In this paper we have attempted to tackle this
problem head on.  Although the test has a few weaknesses, it also has
many unique strengths.  In particular, we believe that it expresses
the essentials of machine intelligence in an elegant and powerful way.
Furthermore, more tractable measures of complexity should lead to
practical tests based on this theoretical model.

\subsection*{Acknowledgments}

This was supported by SNF grant 200020-107616.


\begin{small}

\end{small}

\end{document}